\pdfoutput=1

\documentclass[11pt]{article}

\usepackage{ACL2023}

\usepackage{times}
\usepackage{latexsym}
\usepackage{booktabs}
\usepackage{graphicx}

\usepackage[T1]{fontenc}

\usepackage[utf8]{inputenc}

\usepackage{microtype}

\usepackage{inconsolata}

\title{Rudolf Christoph Eucken at SemEval-2023 Task 4: An Ensemble Approach for Identifying Human Values from Arguments}

\author{Sougata Saha, Rohini Srihari\\
State University of New York at Buffalo\\
Department of Computer Science and Engineering\\
\texttt{\{sougatas, rohini\}@buffalo.edu}}

\begin{document}
\maketitle
\begin{abstract}
The subtle human values we acquire through life experiences govern our thoughts and gets reflected in our speech. It plays an integral part in capturing the essence of our individuality and making it imperative to identify such values in computational systems that mimic human actions. Computational argumentation is a field that deals with the argumentation capabilities of humans and can benefit from identifying such values. Motivated by that, we present an ensemble approach for detecting human values from argument text. Our ensemble comprises three models: (i) An entailment-based model for determining the human values based on their descriptions, (ii) A Roberta-based classifier that predicts the set of human values from an argument. (iii) A Roberta-based classifier to predict a reduced set of human values from an argument. We experiment with different ways of combining the models and report our results. Furthermore, our best combination achieves an overall F1 score of 0.48 on the main test set.

\end{abstract}

\section{Introduction}

Human values \cite{searle2003rationality} play an integral role in determining how and why people react and respond to events in a particular way \cite{schwartz1994there}. We acquire them throughout our lifetime, and they largely govern our actions and manifest through our arguments. They play a significant role in defining our individuality and show a glimpse of our inner self by revealing our belief system. Motivated by why people reason and respond in a certain way, we present computational models capable of automatically identifying such perceivable values from argument text.

Given a set of 20 human value categories, the task \cite{kiesel:2023} comprises classifying the most likely value categories from textual arguments in English. Our approach includes an ensemble comprising three models: (i) An entailment-based model for determining the human values based on their descriptions, (ii) A Roberta-based classifier that predicts the set of human values from an argument. (iii) A Roberta-based classifier to predict a reduced set of human values from an argument. Compared to standard classification-based approaches like models (ii) and (iii), we observe superior results from the entailment-based model. We experiment with different ways of combining the models\footnote{We release our code here: https://github.com/sougata-ub/semeval\_2023\_rudolf\_christoph\_eucken}, attaining an overall F1 score of 0.48 on the main test set, which is an improvement over the baseline of 0.42. 







\section{Background}
We used the standard training, validation and test splits of the shared-task dataset \cite{mirzakhmedova:2023} in English, which is based on the Webis-ArgValues-22 dataset \cite{kiesel:2022}. The dataset comprises 9,324 arguments (5,393 train, 1,996 validation, and 1,935 test) from 6 diverse sources, covering religious texts, political discussions, free-text arguments, newspaper editorials and online democracy platforms. Each argument is manually annotated for 20 human value categories (L2) spanning 54 human values (L1). Given a textual argument and a human value category, the task involves classifying whether an argument draws on that category.

	
	

\section{System Overview}
Our implementation comprises an ensemble of three models, where a primary model performs textual entailment and two secondary models perform classical text classification.

\begin{figure*}[t]
  \centering
  \includegraphics[width=0.91\linewidth]{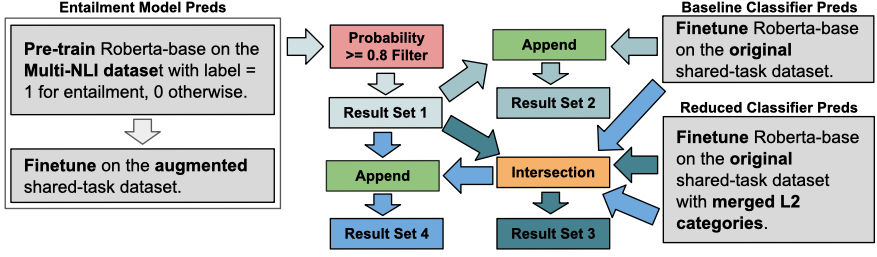}
  \caption{Ensemble Prediction Pipeline.}
  \label{fig:prediction-pipeline}
\end{figure*}

\subsection{Textual Entailment Model}
Although the task constitutes identifying human value categories (L2) from argument text, the dataset also contains finer human values (L1) and their textual descriptions. For example, the value category `Self-direction: thought' represents the values `Be creative', `Be curious' and `Have freedom of thought'. Each value is described by a set of sentences explaining what it means to possess such a human value. For example, `promoting imagination' and `being more creative' are two of the several descriptors for the value `Be creative`. We implement an entailment-based model for identifying the value descriptors that an argument text entails. We transform the dataset conducive for textual entailment by creating argument and value descriptor pairs with its binary L1 label as the target to train our entailment-based model. We report the predictions aggregated at the L2 level.

\begin{figure}[h]
  \centering
  \includegraphics[width=\columnwidth]{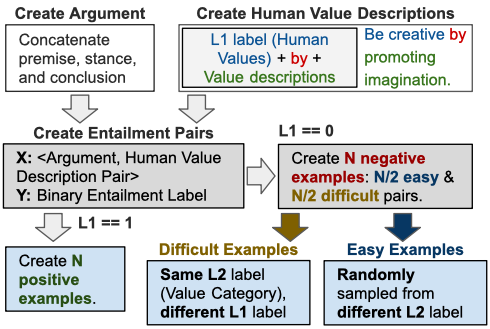}
  \caption{Textual Entailment Dataset Creation Pipeline.}
  \label{fig:entail-data}
\end{figure}

Figure \ref{fig:entail-data} illustrates our entailment dataset creation pipeline. Each example comprises an argument and value description pair and a binary label for entailment. The argument text is created by concatenating the premise, stance, and conclusion. The description text is created by prepending a value descriptor with its L1 label and the keyword `by'. We yield N positive and N negative entailment examples for an argument, where N is the total number of value descriptors across all its positive L1 labels. Further, the N negative examples comprise equal proportions of `easy' and `difficult' samples, where a `difficult' descriptor belongs to the same value category and different value, whereas an `easy' descriptor is randomly sampled from a distinct value category altogether. The entailment dataset comprises 187,058 training and 65,900 validation pairs, which we use to train a Roberta-based \cite{liu2019roberta} entailment model.

We follow a two-phased approach for training the entailment-based model. First, we pre-train Roberta-base on the MultiNLI \cite{multi-nli} dataset. For our purpose, we binarize the labels by assigning 1 to the entailment class and 0 otherwise. We train the model for one epoch using mini-batches of 32 examples and a learning rate of 1e-5. The model is optimized using AdamW \cite{adamw} and achieves a macro F1 score of 0.91 on the combined standard validation splits (validation matched and validation mismatched).
Finally, we fine-tune the resultant model on our curated entailment dataset to engender the final model. We use the same settings as pre-training and achieve a macro F1 score of 0.79 on our validation split.

\subsection{Textual Classification Model}

\begin{table*}
\centering\small%
\setlength{\tabcolsep}{2.5pt}%
\begin{tabular}{@{}ll@{\hspace{10pt}}c@{\hspace{5pt}}cccccccccccccccccccccc@{}}
\toprule
\multicolumn{2}{@{}l}{\bf Test set / Approach} & \bf All & \rotatebox{90}{\bf Self-direction: thought} & \rotatebox{90}{\bf Self-direction: action} & \rotatebox{90}{\bf Stimulation} & \rotatebox{90}{\bf Hedonism} & \rotatebox{90}{\bf Achievement} & \rotatebox{90}{\bf Power: dominance} & \rotatebox{90}{\bf Power: resources} & \rotatebox{90}{\bf Face} & \rotatebox{90}{\bf Security: personal} & \rotatebox{90}{\bf Security: societal} & \rotatebox{90}{\bf Tradition} & \rotatebox{90}{\bf Conformity: rules} & \rotatebox{90}{\bf Conformity: interpersonal} & \rotatebox{90}{\bf Humility} & \rotatebox{90}{\bf Benevolence: caring} & \rotatebox{90}{\bf Benevolence: dependability} & \rotatebox{90}{\bf Universalism: concern} & \rotatebox{90}{\bf Universalism: nature} & \rotatebox{90}{\bf Universalism: tolerance} & \rotatebox{90}{\bf Universalism: objectivity} \\
\midrule
\multicolumn{2}{@{}l}{\emph{Main}} \\
& \textcolor{gray}{Best per category} & \textcolor{gray}{.59} & \textcolor{gray}{.61} & \textcolor{gray}{.71} & \textcolor{gray}{.39} & \textcolor{gray}{.39} & \textcolor{gray}{.66} & \textcolor{gray}{.50} & \textcolor{gray}{.57} & \textcolor{gray}{.39} & \textcolor{gray}{.80} & \textcolor{gray}{.68} & \textcolor{gray}{.65} & \textcolor{gray}{.61} & \textcolor{gray}{.69} & \textcolor{gray}{.39} & \textcolor{gray}{.60} & \textcolor{gray}{.43} & \textcolor{gray}{.78} & \textcolor{gray}{.87} & \textcolor{gray}{.46} & \textcolor{gray}{.58} \\
& \textcolor{gray}{Best approach} & \textcolor{gray}{.56} & \textcolor{gray}{.57} & \textcolor{gray}{.71} & \textcolor{gray}{.32} & \textcolor{gray}{.25} & \textcolor{gray}{.66} & \textcolor{gray}{.47} & \textcolor{gray}{.53} & \textcolor{gray}{.38} & \textcolor{gray}{.76} & \textcolor{gray}{.64} & \textcolor{gray}{.63} & \textcolor{gray}{.60} & \textcolor{gray}{.65} & \textcolor{gray}{.32} & \textcolor{gray}{.57} & \textcolor{gray}{.43} & \textcolor{gray}{.73} & \textcolor{gray}{.82} & \textcolor{gray}{.46} & \textcolor{gray}{.52} \\
& \textcolor{gray}{BERT} & \textcolor{gray}{.42} & \textcolor{gray}{.44} & \textcolor{gray}{.55} & \textcolor{gray}{.05} & \textcolor{gray}{.20} & \textcolor{gray}{.56} & \textcolor{gray}{.29} & \textcolor{gray}{.44} & \textcolor{gray}{.13} & \textcolor{gray}{.74} & \textcolor{gray}{.59} & \textcolor{gray}{.43} & \textcolor{gray}{.47} & \textcolor{gray}{.23} & \textcolor{gray}{.07} & \textcolor{gray}{.46} & \textcolor{gray}{.14} & \textcolor{gray}{.67} & \textcolor{gray}{.71} & \textcolor{gray}{.32} & \textcolor{gray}{.33} \\
& \textcolor{gray}{1-Baseline} & \textcolor{gray}{.26} & \textcolor{gray}{.17} & \textcolor{gray}{.40} & \textcolor{gray}{.09} & \textcolor{gray}{.03} & \textcolor{gray}{.41} & \textcolor{gray}{.13} & \textcolor{gray}{.12} & \textcolor{gray}{.12} & \textcolor{gray}{.51} & \textcolor{gray}{.40} & \textcolor{gray}{.19} & \textcolor{gray}{.31} & \textcolor{gray}{.07} & \textcolor{gray}{.09} & \textcolor{gray}{.35} & \textcolor{gray}{.19} & \textcolor{gray}{.54} & \textcolor{gray}{.17} & \textcolor{gray}{.22} & \textcolor{gray}{.46} \\
& Result Set 1 & .47 & .40 & .60 & .20 & .23 & .60 & .41 & .49 & .25 & .70 & .57 & .61 & .50 & .46 & .14 & .49 & .24 & .71 & .80 & .20 & .37 \\
& Result Set 2 & .48 & .39 & .62 & .20 & .23 & .61 & .43 & .48 & .25 & .70 & .57 & .62 & .54 & .46 & .13 & .49 & .24 & .71 & .80 & .26 & .43 \\
& Result Set 3 & .33 & .38 & .36 & .20 & .07 & .26 & .07 & .37 & .15 & .51 & .11 & .09 & .19 & .09 & .13 & .49 & .23 & .71 & .80 & .20 & .36 \\
& Result Set 4 & .48 & .40 & .60 & .20 & .23 & .60 & .41 & .48 & .25 & .70 & .57 & .61 & .51 & .46 & .13 & .49 & .24 & .71 & .80 & .26 & .42 \\
\bottomrule
\end{tabular}
\caption{Achieved F$_1$-score of team rudolf-christoph-eucken per test dataset, from macro-precision and macro-recall (All) and for each of the 20~value categories. Approaches in gray are shown for comparison: an ensemble using the best participant approach for each individual category; the best participant approach; and the organizer's BERT and 1-Baseline.}
\label{table-results}
\end{table*}

We create argument texts as previously discussed, and train two variants of Roberta multi-label classifiers for directly predicting the perceivable value categories: (i) \textbf{Baseline Classifier:} We use the original 20 value categories as the target variable. (ii) \textbf{Reduced Classifier:} We split value category labels using `:' as the delimiter and combine the value categories starting with the same prefix, yielding a set of 12 classes. For example, the classes `Self-direction: thought' and `Self-direction: action' are combined into one class as `Self-direction'. Leveraging Roberta-base as the base model, both classifiers uses a linear layer for predicting the target classes from the pooler output. We train both models for 30 epochs with mini-batches of 64 examples and a learning rate of 2e-5. We use AdamW as the optimizer and stop training if the validation loss doesn't improve for 4 epochs.

	
	
	
	
	

\section{Experimental Setup}

Using each of the three models independently, we perform predictions on the standard shared-task validation and test sets. Illustrated in Figure \ref{fig:prediction-pipeline}, we combine the independent results using the following four schemes to generate the final set of predictions: (i) \textbf{Result Set 1:} Using the entailment model, we predict on the test set. We compare each example with all the value descriptors to check for entailment and filter out predictions less than probability 0.8. We report the value categories of the final value descriptors. (ii) \textbf{Result Set 2:} To increase the system's recall, we append the test set predictions from the baseline classifier with the result set 1. (iii) \textbf{Result Set 3:} Contrary to result set 2, here we try increasing the system's precision by reporting only the intersecting labels between the result set 1 and the reduced classifier predictions. (iv) \textbf{Result Set 4:} For each test set example, we first compute the intersecting labels from the baseline and reduced classifier predictions and append with the result set 1. Across all runs, we use TIRA \cite{froebe:2023} as the evaluation platform. We compute and compare overall and value category-wise F1, precision, and recall for the standard validation and test sets.

	
	
	
	
	

\section{Results}


We report our test set results in Table \ref{table-results}. Although our system falls short of the best approach, it performs significantly better than BERT and 1-Baseline, achieving a leaderboard position of 17th out of 41 teams in TIRA. As depicted in Table \ref{tab:model-comparison}, compared to only classification-based systems (Baseline and Reduced), the entailment-based approach independently yields superior validation results. Hence, we used the entailment-based model output as our base result set (Result Set 1) and leveraged the classification-based model outputs to improve the entailment-based results. As indicated in Table \ref{table-results}, we observe that appending the classification-based results enhances the performance on the test set (in Result Set 2 and 4), whereas filtering out the entailment-based results based on the classification results (in Result Set 3) hampers performance. 

\begin{table}[h]
\centering
\resizebox{0.85\columnwidth}{!}{%
\begin{tabular}{lccc}
\hline
\multicolumn{1}{c}{\textbf{Model}} & \textbf{F1} & \textbf{Precision} & \textbf{Recall} \\ \hline
Textual Entailment                  & 0.49        & 0.44               & 0.56            \\
Baseline Classifier                 & 0.39        & 0.56               & 0.30            \\
Reduced Classifier                  & 0.26        & 0.17               & 0.53            \\ \hline
\end{tabular}%
}
\caption{Overall model comparison on validation set.}
\label{tab:model-comparison}
\end{table}

We reason that since models like Roberta use large data samples for training, they learn to encode the semantics better: They can discern the meaning of an argument and associate it with the value descriptions. Furthermore, Roberta is pre-trained using the next-sentence prediction task and can associate relationships between sentence pairs. On the contrary, since classification-based approaches require learning domain-specific relationships, they are bottlenecked by the volume of the available training data. Hence, an ensemble approach such as Result Set 2 and 4 leverages the best of both scenarios and yields the best results.

	
	
	


\section{Conclusion}
Computational models for automatic detection of human values from arguments is a nascent yet crucial research direction for enabling purposeful argumentation systems. Here we present an ensemble comprising entailment and classification-based models for detecting human values from argument text. We experiment with different ways of combining the model outputs and attain an overall F1 score of 0.48 on the main test set, which significantly improves the baseline. 


\bibliography{custom}
\bibliographystyle{acl_natbib}



\end{document}